\documentclass[a4paper, 10pt, conference]{IEEEtran}
\IEEEoverridecommandlockouts

\usepackage{cite}
\usepackage{amsmath,amssymb,amsfonts}
\usepackage{algorithmic}
\usepackage{algorithm}
\usepackage{graphicx}
\usepackage{textcomp}
\usepackage{xcolor}
\ifCLASSOPTIONcompsoc
    \usepackage[caption=false, font=normalsize, labelfont=sf, textfont=sf]{subfig}
\else
\usepackage[caption=false, font=footnotesize]{subfig}
\fi

\def\BibTeX{{\rm B\kern-.05em{\sc i\kern-.025em b}\kern-.08em
    T\kern-.1667em\lower.7ex\hbox{E}\kern-.125emX}}

\begin{document}

\title{Autonomous Warehouse Robot using Deep Q-Learning\\

\thanks{\textsuperscript{1} Equal Contribution.

\textsuperscript{2} Undergraduate student.

\textsuperscript{3} Assistant Professor, IEEE Member.}
}

\author{\IEEEauthorblockN{Ismot Sadik Peyas\textsuperscript{1, 2}, Zahid Hasan\textsuperscript{1, 2}, Md. Rafat Rahman Tushar\textsuperscript{1, 2}, Al Musabbir\textsuperscript{2}, Raisa Mehjabin Azni\textsuperscript{2},\\
Shahnewaz Siddique\textsuperscript{3}}
\IEEEauthorblockA{\textit{Department of Electrical and Computer Engineering} \\
\textit{North South University}\\
Dhaka, Bangladesh \\
\{ismot.peyas, zahid.hasan04, rafat.tushar, al.musabbir, raisa.azni, shahnewaz.siddique\}@northsouth.edu}
}

\maketitle


\begin{abstract}

In warehouses, specialized agents need to navigate, avoid obstacles and maximize the use of space in the warehouse environment. Due to the unpredictability of these environments, reinforcement learning approaches can be applied to complete these tasks. In this paper, we propose using Deep Reinforcement Learning (DRL) to address the robot navigation and obstacle avoidance problem and traditional Q-learning with minor variations to maximize the use of space for product placement. We first investigate the problem for the single robot case. Next, based on the single robot model, we extend our system to the multi-robot case. We use a strategic variation of Q-tables to perform multi-agent Q-learning. We successfully test the performance of our model in a 2D simulation environment for both the single and multi-robot cases.\\
\end{abstract}



\section{Introduction}
The global warehouse robotics market is predicted to grow at a CAGR (Compound Annual Growth Rate) of 14.0\%, from USD (United States Dollar) 4.7 billion in 2021 to USD 9.1 billion by 2026~\cite{warehouse_robot_market}. According to Dubois and Hamilton~\cite{team_who_created} the need for warehouse robots is growing, and is expected to expand. In 2017, these warehouse robots assisted in the picking and packing of goods worth USD 394.8 billion. 

The impact of COVID-19 on the market resulted in a massive increase in demand for warehouse robots~\cite{warehouse_robot_market}. The pandemic's supply chain disruption is impacting the market severely. Additionally, due to lockdown and travel restrictions, companies are not able to get the necessary workforce for their operations. Various warehouse operations, such as transportation, picking and placing, packaging, palletizing, and de-palletizing, are automated using warehouse robotics. The deployment of warehouse robots minimizes the need for human interaction and improves warehouse operations efficiency. Warehouse robots are used in a variety of fields such as online shopping, automotive, electrical, electronics, food and beverage, and pharmaceuticals to name a few. 

For a sustainable supply chain system, these operations must be executed fast and efficiently. Both autonomous Unmanned Ground Vehicles (UGV) and Unmanned Aerial Vehicles (UAV)  can be very efficient in such scenarios. Such warehouse agents can be utilized with autonomous algorithms to conduct operations that are challenging for human operators at low operating costs. Warehouse operations involve receiving, shipping and storing. Stacking loaded pallets in warehouses and storage facilities are critical for preventing accidents. Poorly stacked loaded pallets pose a severe risk to employee safety and can cause significant product damage and increase the total cost of business. Also, in many cases maintaining the health and safety of a human workforce becomes costlier than maintaining a fleet of robots.

The warehouse environment varies from place to place based on their construction and architectural design. Therefore, in many cases, a precise mathematical model of the underlying environment is unavailable or ambiguous. So, it is vital to build an efficient and accurate model to address these complicated tasks without human interference. Moreover, the search environment can change unexpectedly, and the objects can be placed anywhere in the warehouse. Hence, the agent's interaction with the environment should be autonomous, and the agent must have the capability to make decisions for itself.

On such occasions, reinforcement learning (RL)~\cite{rl-intro} proposes a unique approach to solve these issues. RL does not require any prior knowledge of the environment. Agents based on RL algorithms can navigate the environment autonomously without any explicit model of the environment. Rather, the RL agent frequently interacts with the environment and receives negative or positive rewards based on a predefined reward function. Through this process, it learns to function in an entirely new environment.

Our agent function consists of three major components: (1) autonomous navigation, (2) stacking products optimally, and (3) obstacle avoidance. The autonomous navigation and obstacle avoidance feature is based on Deep Q-learning. The agent has a set of forward, backward, left, and right actions to navigate and avoid collisions in the warehouse environment. The robot finds the maximum available space in the warehouse and then moves the product using the shortest path available to the destination point. The destination space is updated as soon as the product is place in the destination point (maximum available space). Discovering the maximum available space is implemented with the Q-learning algorithm. 

Our system is first developed for the single robot case. Later, a multi robot system is also developed to operate in the warehouse environment. In the multi-agent system, all agents aim to maximize their cumulative reward. When an agent collides with an obstacle or another agent, their reward is deducted by a certain amount.

\section{Related Work}



Reinforcement learning is not widely used in warehouse robotics research. In warehouse operations, path finding and obstacle avoidance are challenging. The most popular approaches employed in path computing to meet this difficulty are deterministic, heuristic-based algorithms~\cite{heuristic}. \cite{heuristic} compares and contrasts static algorithms (such as A*), re-planning algorithms (such as D*), anytime algorithms (such as ARA*), and anytime re-planning algorithms (such as AD*). Classical algorithms generate path planning for known static environments. In path planning, states are agent locations and transitions between states are actions the agent can do, each with a cost~\cite{heuristic}. Later these are expanded and blended to work in a partially known or dynamic environment.

A path planning algorithm is required for the mobile robot to operate autonomously throughout the warehouse~\cite{collisoin_free_path}. For the mobile robot, this path planning algorithm generates a collision-free path from the start point to the goal point. The location of all the shelves and the open space must be known to the algorithm in order for it to complete this task. In our study, we have used Reinforcement learning, which does not require this information. Once the algorithm has been given the start and destination points, it will evaluate all four nearby grids to see if they are shelves or free space. In works such as ~\cite{collisoin_free_path} the closest euclidean distance between all nearby free space grids and the objective point is considered after identifying the neighboring free space grids, whereas our agent is reward driven. This process is repeated until the distance between the goal and the present point reaches zero.

Reinforcement learning algorithms have already been utilized to develop algorithms for an autonomous aerial vehicle that can rescue missing people or livestock~\cite{winderness}. \cite{winderness}  used Deep Q learning for robot navigation. They used a cyclic approach of three tasks: Region Exploration, Target Search, and Target Selection. The DQN architecture explicitly separates the representation of state values and state-dependent action advantages via two separate streams.

In~\cite{rough_terrain}, the authors developed and trained a Deep Reinforcement Learning (DRL) network to determine a series of local navigation actions for a mobile robot to execute. The onboard sensors on the robot provided the sensory data.The results showed that using the DRL method the robot could successfully navigate in an environment towards the target goal location when the rough terrain is unknown. 

A system for fast autonomy on a quadrotor platform showed its capabilities and robustness in high-speed navigation tasks~\cite{gps_quadrotor}. As the speed rises, state estimation, planning, and control difficulties increase significantly. These issues are rectified based on the existing methods and demonstrate the whole system in various environments~\cite{gps_quadrotor}. To avoid the obstacle, our model uses the deep learning method and object detection is crucial.

\cite{object_detection} presents a review of deep learning-based object detection frameworks. It initially focuses on typical generic object detection architectures and some modifications and valuable tricks to improve detection performance. As distinct particular detection tasks show various characteristics,~\cite{object_detection} briefly survey numerous specific tasks, including salient object detection, face detection, and pedestrian detection. Experimental studies are also given to distinguish various methods. Finally, some promising directions and tasks are provided to serve as guidelines for future work in both object detection and relevant neural network-based learning systems.


\section{Deep Q-Learning for Warehouse Agents}

\subsection{Deep Q-learning}\label{sec-q-learning}
Any discrete, stochastic environment can be described as Markov Decision Process (MVP). MVP is the mathematical formulation of intelligent decision-making processes. According to MVP, an actor or agent, given an environment, $E$, performs a task or takes action at time $t$ and transits into a new state $s_{t+1}$ of that environment at a time $(t+1)$. This can be written as, 

\begin{equation}
    \label{eq1}
    f(S_t, A_t) = R_{t+1}
\end{equation}

The reward can further be described as a discounted reward, where the agent takes action following a policy, which provides the agent with the future discounted reward of this present action. The discounted reward can be formulated as,

\begin{equation}
    \label{eq2}
    G_t=R_{t+1}+\gamma R_{t+2}+\gamma^2 R_{t+3}+...
\end{equation}

Here, $\gamma$ is the discounted factor which is between 0 and 1. The maximum discounted reward depends on the optimal state-action value pair followed by the policy. Q-learning is based on this MVP paradigm. By following this process, the optimal q-function can be written as, 

\begin{equation}
    \label{eq3}
    q^*(s,a) = max~q(s,a)
\end{equation}

According to this q-function, the policy should choose the highest q-value to get the highest future overall reward. To get the optimal q-value, the Bellman Equation~\cite{bellman} must be satisfied. Therefore, we can write,

\begin{figure*}[!t]
\centering
\subfloat[Navigation and obstacle avoidance system\label{subfig-env-nav-obs}]
{\includegraphics[trim={3cm 1.2cm 3cm 1cm},clip,width=0.16\linewidth, height=2.6cm]{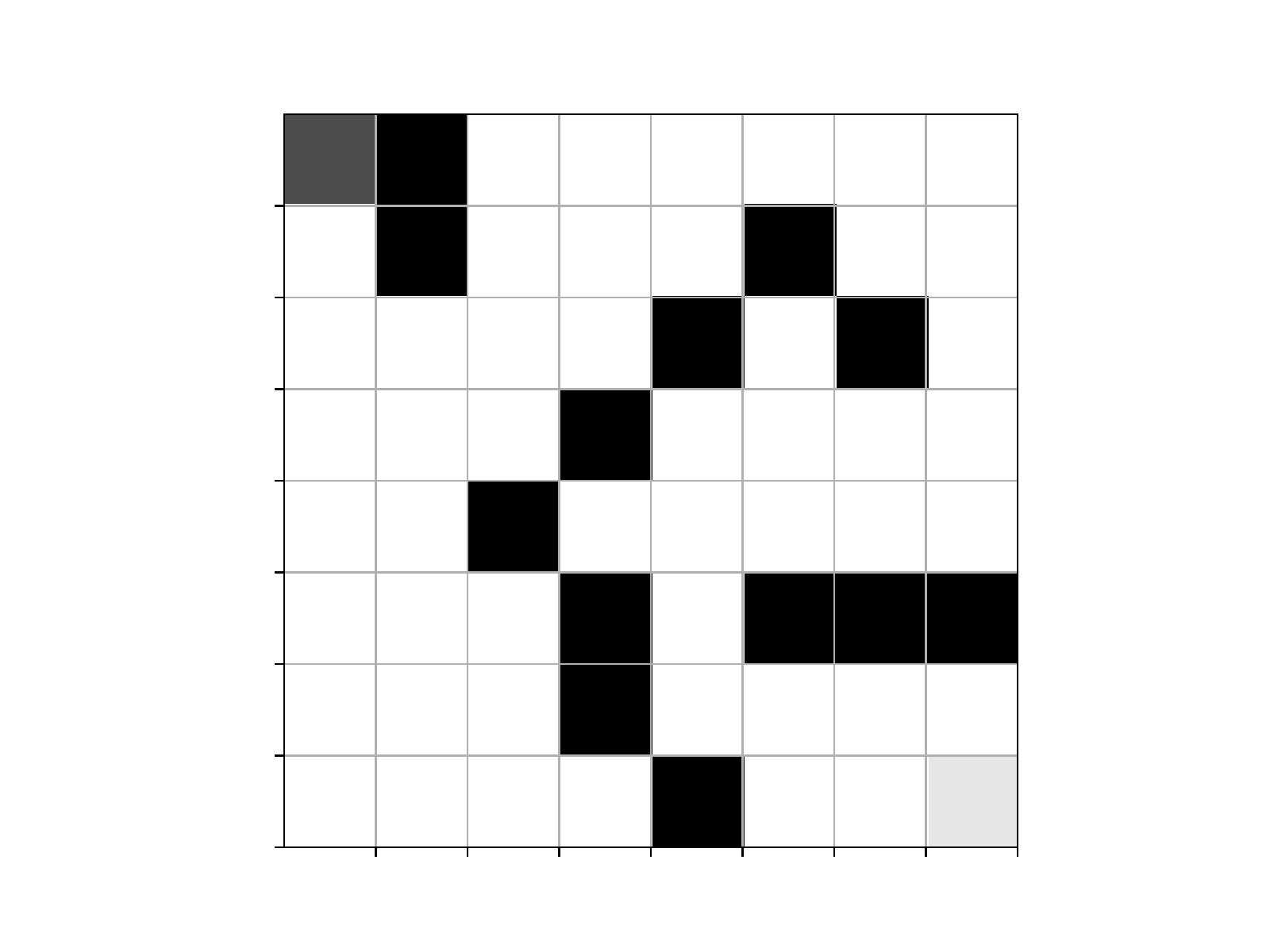}}
\hfil
\subfloat[Maximum available space finding\label{subfig-env-max-space}]
{\includegraphics[trim={3cm 1cm 3cm 1cm},clip,width=0.16\linewidth, height=2.5cm]{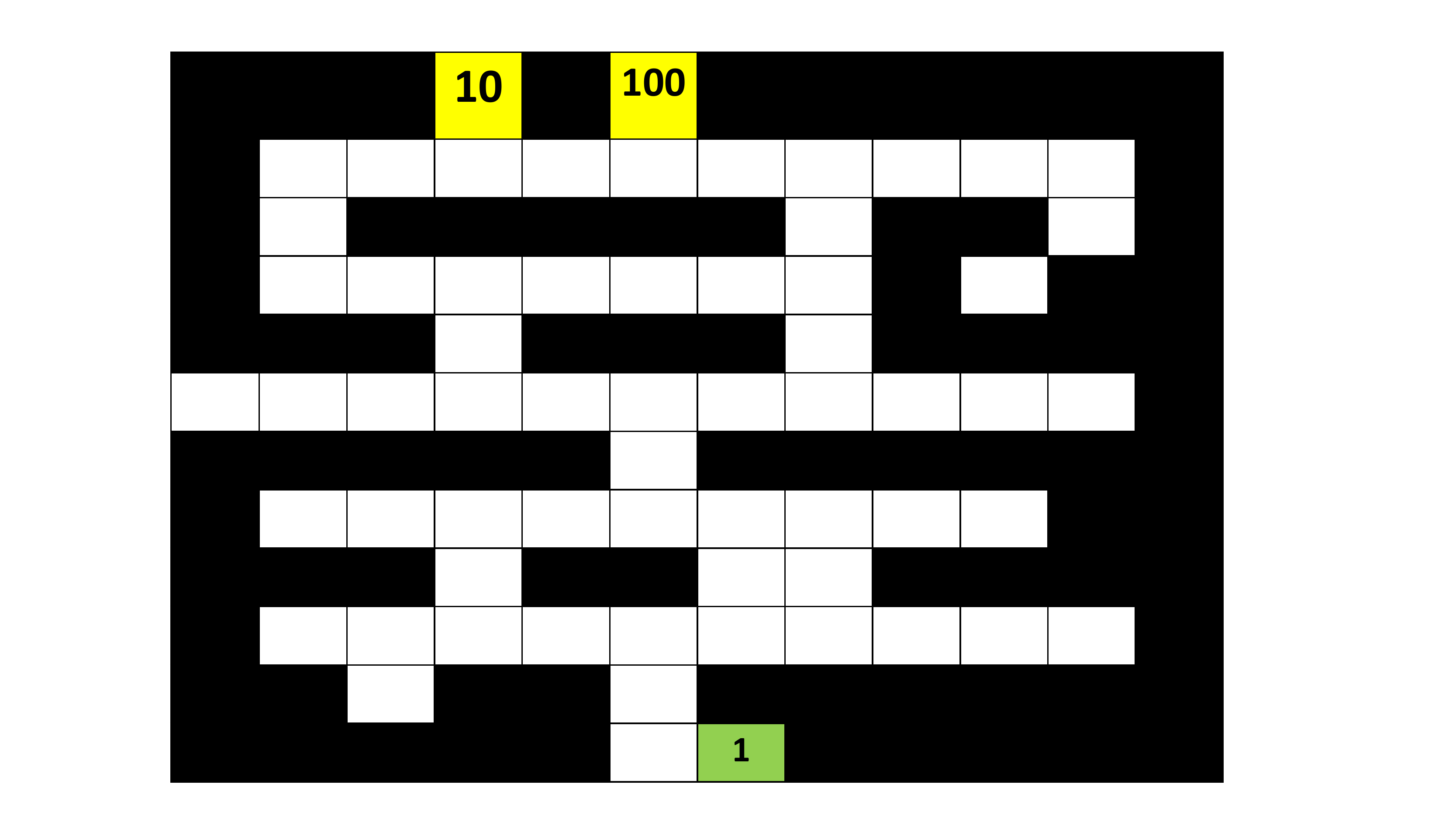}}
\hfil
\subfloat[Multi-agent environment\label{subfig-env-multi}]
{\includegraphics[width=0.16\linewidth, height=2.5cm]{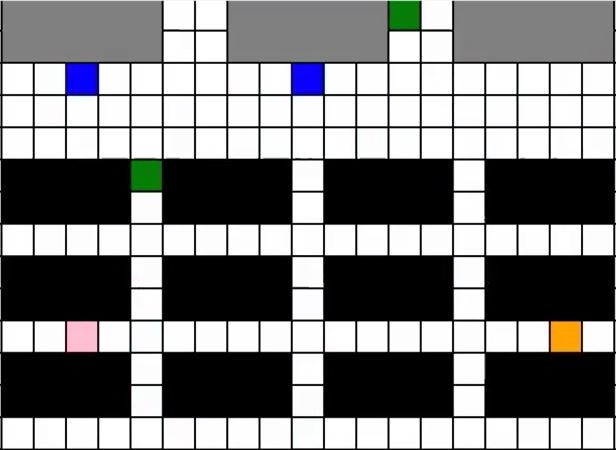}}
\caption{Environment design}
\label{env_all}
\end{figure*}

\begin{equation}
    \label{eq4}
    q^*(s,a)=E[R_{t+1}+\gamma max~q^*(s',a')]
\end{equation}

This equation means that the optimal q-value function for a given state-action pair $(s,a)$ will be the expected reward $R_{t+1}$ after taking that action plus the maximum discounted reward by following that optimal policy for the future state-action pair $(s',a')$. To find this q* value, sometimes a linear function approximator is used if the state space is simple. But in a complex environment, a non-linear function approximator, like the neural network, is used to approximate the optimal q-value.

\subsection{Navigation and Obstacle Avoidance}\label{sec-nav-obs}
\begin{algorithm}[!t]
\caption{Navigation and Obstacle Avoidance with Deep Q-learning}
\label{algo-nav-obs}
\begin{algorithmic}

\REQUIRE Initialize the warehouse 2D environment, replay memory buffer $E$ with capacity $C_E$, q-function approximator neural network $Q_n$ with random weights, exploration probability $\epsilon$, discount factor $\gamma$
\FOR{episode = 1 to M}
    \STATE Get starting state $s_0$
    \WHILE{episode not terminated}
        \STATE Take random $\rho$ value between 0 and 1
        \IF{$\epsilon > \rho$}
            \STATE Take random action $a_t$ from action space
        \ELSE
            \STATE Action $a_t = argmax_a Q(s_t, a)$
        \ENDIF
        \STATE Decay exploration probability $\gamma$
        \STATE Perform action $a_t$ then get reward $r_t$ and next state $s_{t+1}$
        \STATE Store experience tuple $e_t = (s_t, a_t, r_t, s_{t+1})$ in replay memory $E$ in FIFO order
        \STATE Take  sample minibatch $(s_k, a_k, r_k, s_{k+1})$ from $E$
        \IF{episode terminates at $s_{k+1}$}
            \STATE Set $y_k = r_k$
        \ELSE
            \STATE Set $y_k = r_k + \gamma max_a Q(s_{k+1},a)$
        \ENDIF
        \STATE Calculate loss function $loss = (y_k - Q(s_k, a_k))^2$
        \STATE Execute optimization algorithm Adam~\cite{adam}  on loss function to update neural network $Q$ for back-propagation
        \STATE Set $s_t = s_{t+1}$
    \ENDWHILE
\ENDFOR


\end{algorithmic}
\end{algorithm}

\begin{figure}[!b]
    \centering
    \includegraphics[width=0.6\linewidth, height=2.8cm]{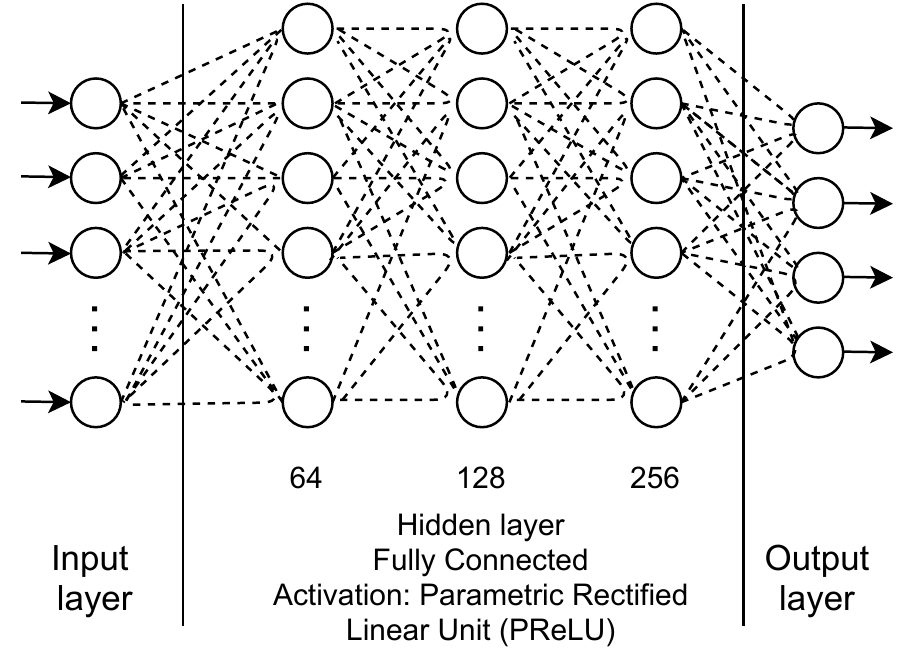}
    \caption{Neural network structure}
    \label{fig-neural-net}
\end{figure}

When constructing a warehouse environment agent, we first structure the warehouse upper-view as a 2D map divided into $8\times8$ equal regions. For simplicity, we assumed that our warehouse would only contain boxes of the same length and width. The warehouse agent has access to the upper view of the environment. That means we can train the agent on this 2D map array. We define the starting point, $s = (x_0,y_0)$, and the map's destination point, $d = (x_d,y_d)$. The warehouse 2D cells have values of one (1) and zero (0). The zero represents there is an obstacle, and one represents the open path. The visual representation of this warehouse environment is shown in Fig.~\ref{env_all}\subref{subfig-env-nav-obs}. Here the starting point is the upper-left cell, and the destination cell for the agent is the lower-right cell. The black boxes are the walls or obstacles, and white boxes are the allowed moving paths. The agent can move freely with four action spaces: front, back, left, and right. The reward mechanism for the agent is simple, which is shown in Table~\ref{reward-nav-table}. We design a simple neural network, which is used as a function approximator for q-values. The architecture of the neural network is given in Fig.~\ref{fig-neural-net}. We have trained our model up to 500 epochs, and if the agent can reach the destination before 500 epochs without hitting any walls, we terminate the training process. The whole process can be observed by looking at Algorithm~\ref{algo-nav-obs}. The hyper-parameters used in this training process are shown in Table~\ref{table-hyper}.

\begin{table}[!t]
\renewcommand{\arraystretch}{1.3}
\caption{Reward Mechanism for Training}
\centering
\begin{tabular}{ccc}
\hline
\textbf{Moves}&\textbf{Rewards}&\textbf{Results} \\
Agent hit with wall/obstacle & -1 & End of an episode \\
Agent in the free-way & 0 & Continue the episode \\
Agent reaches the destination & +1 & End of an episode \\
\hline
\end{tabular}
\label{reward-nav-table}
\end{table}

\begin{table}[!t]
\renewcommand{\arraystretch}{1.3}
\caption{List of Hyperparameters}
\centering
\begin{tabular}{ccc}
\hline
\textbf{Hyperparameter}&\textbf{Value}&\textbf{Description} \\
Discount Factor & 0.90 & $\gamma$-value in max Q-function \\
Initial Epsilon	& 1.0 & Exploration epsilon initial value \\
Final Epsilon & 0.1 & Exploration final epsilon value \\
Batch size & 32 & Mini batch from replay memory \\
Learning Rate & 0.0025 & Learning rate for Adam optimizer \\
Experience Replay & 1000 & Capacity of experience replay \\
Memory          &       & memory \\
\hline
\end{tabular}
\label{table-hyper}
\end{table}

\subsection{Finding the Maximum Available Space for Storing}\label{sec-max-space}

\begin{algorithm}[!t]
\caption{Maximum Available Space with Q-learning}
\label{algo-max-space}
\begin{algorithmic}

\REQUIRE Initialize the warehouse 2D environment, Q table $Q$, exploration probability $\epsilon$, discount factor $\gamma$, q-value updating factor $\alpha$
\FOR{episode = 1 to M}
    \STATE Initialize a random process and get the initial state information $s_0$
    \WHILE{episode not terminated}
        \STATE Take random $\rho$ value between 0 and 1
        \IF{$\epsilon > \rho$}
            \STATE Take random action $a_t$ from action space
        \ELSE
            \STATE Action $a_t = arg\max_a Q(s_t, a_t)$
        \ENDIF
        \STATE Decay exploration probability $\epsilon$
        \STATE Execute action $a_t$ then observe reward $r_t$ and next state $s_{t+1}$
        \STATE Get Q-predict $Q_p = Q(s_t, a_t)$ from Q-table
        \IF{episode terminates at $s_{t+1}$}
            \STATE Target $Q_T = r_t$
        \ELSE
            \STATE Target $Q_T = r_t + \gamma max_a Q(s_{t+1})$
        \ENDIF
        \STATE Update Q-table $Q(s_t, a_t) \mathrel{+}= \alpha (Q_T - Q_P)$
        \STATE Decay updating factor $\alpha$
    \ENDWHILE
\ENDFOR

\end{algorithmic}
\end{algorithm}

We developed a slightly different environment for this training process. Because this time, the agent has to know each cell’s occupied and available space, the warehouse environment must contain that information. The visual design of this environment is shown in Fig.~\ref{env_all}\subref{subfig-env-max-space}. The modified 2D view of the environment has cells containing five different values. The cells' values and their representations are shown in Table~\ref{env-cell-value-b}. The goal for the agent is to learn the shortest possible path to reach the cell that has the most available space. Moreover, the agent has to learn to avoid any obstacle while reaching the optimal destination point. After arriving at the optimal destination, which is 100 in our environment, the available space for that cell is updated. For example, when the object reaches the maximum available space cell, which is 100, the available space for that cell becomes 99. We developed a $12\times12$ 2D map array for training this model. We used the Q-learning algorithm for training our agent to navigate and identify the optimal path and destination through the warehouse environment. Through exploration, our agent can get to know the best possible action that can be taken given a state. The mathematical explanation of Q-learning can be found in Section~\ref{sec-q-learning}. For policy or action-selection strategy, we employed the $\epsilon$-greedy~\cite{epsilon-greedy} approach during training. Equation~\ref{eq4} shows the updating process of optimal Q-function. In section~\ref{sec-q-learning}, we describe that for q-function, sometimes a linear function approximator is used. In this model, we used a vector-based q-table for storing and retrieving the updated q-values. Algorithm~\ref{algo-max-space} contains the detailed implementation of our model.

\begin{table}[!b]
\renewcommand{\arraystretch}{1.3}
\caption{Reward and Cell's Value Representation}
\centering
\begin{tabular}{cc}
\hline
\textbf{Values}&\textbf{Representation} \\
-100 & Wall or obstacle \\
-1 & Open Path \\
100 & One hundred available space \\
10 & Ten available space \\
1 & The object to be stored \\
\hline
\end{tabular}
\label{env-cell-value-b}
\end{table}

\subsection{Multi-agent Exploration}\label{sec-multi-agent}

\begin{algorithm}[!ht]
\caption{Multi-agent Q-learning with Q-tables}
\label{algo-multi}
\begin{algorithmic}

\REQUIRE Initialize the warehouse 2D environment, Q tables $Q_1, Q_2, ..., Q_n$ for n number of agents, exploration probability $\epsilon$, discount factor $\gamma$, Q-value updating factor $\alpha$
\FOR{episode = 1 to M}
    \STATE Initialize a random process and get the initial state information $s_0$
    \WHILE{episode not terminated}
        \FOR{agent i = 1 to n}
            \STATE With probability $\epsilon(i)$ select a random action $a_t(i)$; otherwise, select best available action from q-table
            \STATE Decay exploration probability $\epsilon(i)$
            \STATE Execute action $a_t(i)$ then observe reward $r_t(i)$ and next state $s_{t+1}(i)$
            \STATE Get Q-predict $Q_p = Q(s_t(i), a_t(i))$ from Q-table
            \IF{episode terminates at $s_{t+1}$}
                \STATE Target $Q_T = r_t(i)$
            \ELSE
                \STATE Target $Q_T = r_t(i) + \gamma max_a Q(s_{t+1}(i))$
            \ENDIF
            \STATE Update Q-table $Q(s_t(i), a_t(i)) \mathrel{+}= \alpha (Q_T - Q_P)$
            \STATE Decay updating factor $\alpha$
        \ENDFOR
    \ENDWHILE
\ENDFOR

\end{algorithmic}
\end{algorithm}

The optimal system for the warehouse problem will be a multi-agent environment where more than one agent will interact with the warehouse environment cooperatively. We designed a multi-agent model for our warehouse environment where multiple autonomous actors can store and transport. Fig.~\ref{env_all}\subref{subfig-env-multi} displays the visual representation of multi-agent environment. The two blue boxes act as two agents, while the green boxes represent human workers. The orange and pink boxes are the destination points, the black boxes are the obstacles, and the rest white areas are the free-moving path for the agents. We performed multi-agent Q-learning with a strategic variation of Q-tables. We initially create Q-tables for each agent and use these tables to store q-values for state-action pairs during training. We train our agents on these Q-tables containing q-values for every possible optimal navigation from the initial position to the destination in the warehouse environment. The q-values are stored and updated in the Q-tables by the factor of $\alpha$, which we call the q-value update factor. This variable is used to control the impact of updating and storing q-values. Primarily, q-values are updated with much higher impact or higher factors in the Q-tables. As time passes, the q-value updating impact is reduced by using this q-value update factor $\alpha$. Initially, we set $\alpha$ value 0.03. This value decays by the factor 0.002 times the current episode until it reaches 0.001. The idea of the q-value updating factor is that primarily our q-values in Q-tables contain values that can be sometimes noisy or wrong, and more impactful updates are needed to those values if any optimal state-action values are observed. But, after some training, the q-values in Q-tables often are more accurate, and it may cause harm to make significant changes to those accurate q-values. So, as time passes, the impact of updating the q-values needs to be reduced by a factor which is $\alpha$. After successful training, given a state, the agents can predict the optimal action to be taken by exploring the respective Q-tables for each agent. The optimal action refers to the action which provides the maximum reward among all possible actions that can be taken with a given state. The detailed procedure of our multi-agent model is provided in the Algorithm~\ref{algo-multi}. Here, we have designed a warehouse environment with two autonomous agents, two moving humans, and some obstacles. Our autonomous agents have successfully learned optimal strategies for navigating and reaching a destination without collision with the other agent, the obstacles, and the humans.


\section{Results and Analysis}

\begin{figure}[!t]
\centering
\subfloat[Loss vs. epoch graph\label{res-nav-1}]
{\includegraphics[width=0.47\linewidth]{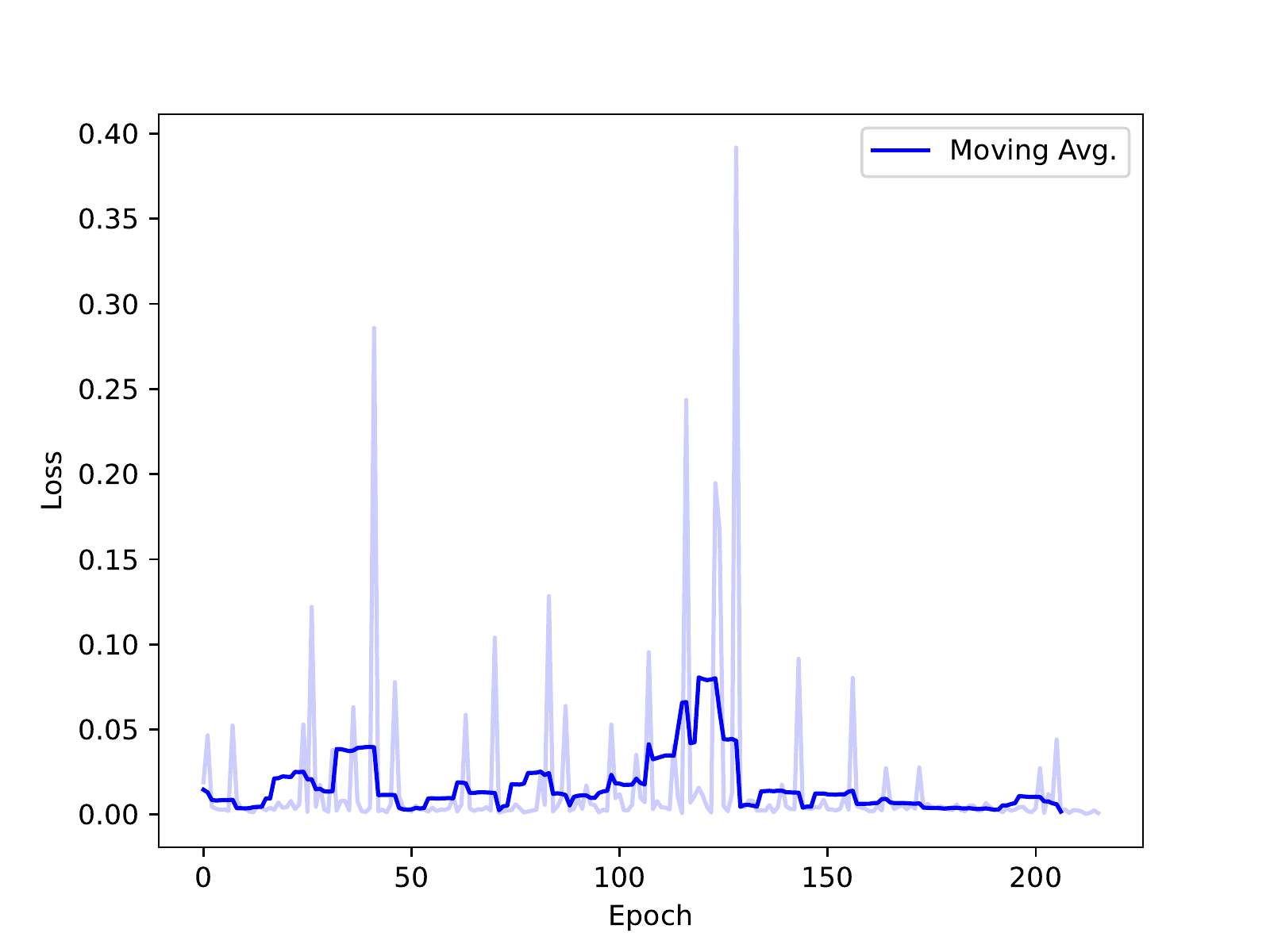}}
\hfil
\subfloat[Win rate vs. epoch graph\label{res-nav-2}]
{\includegraphics[width=0.47\linewidth]{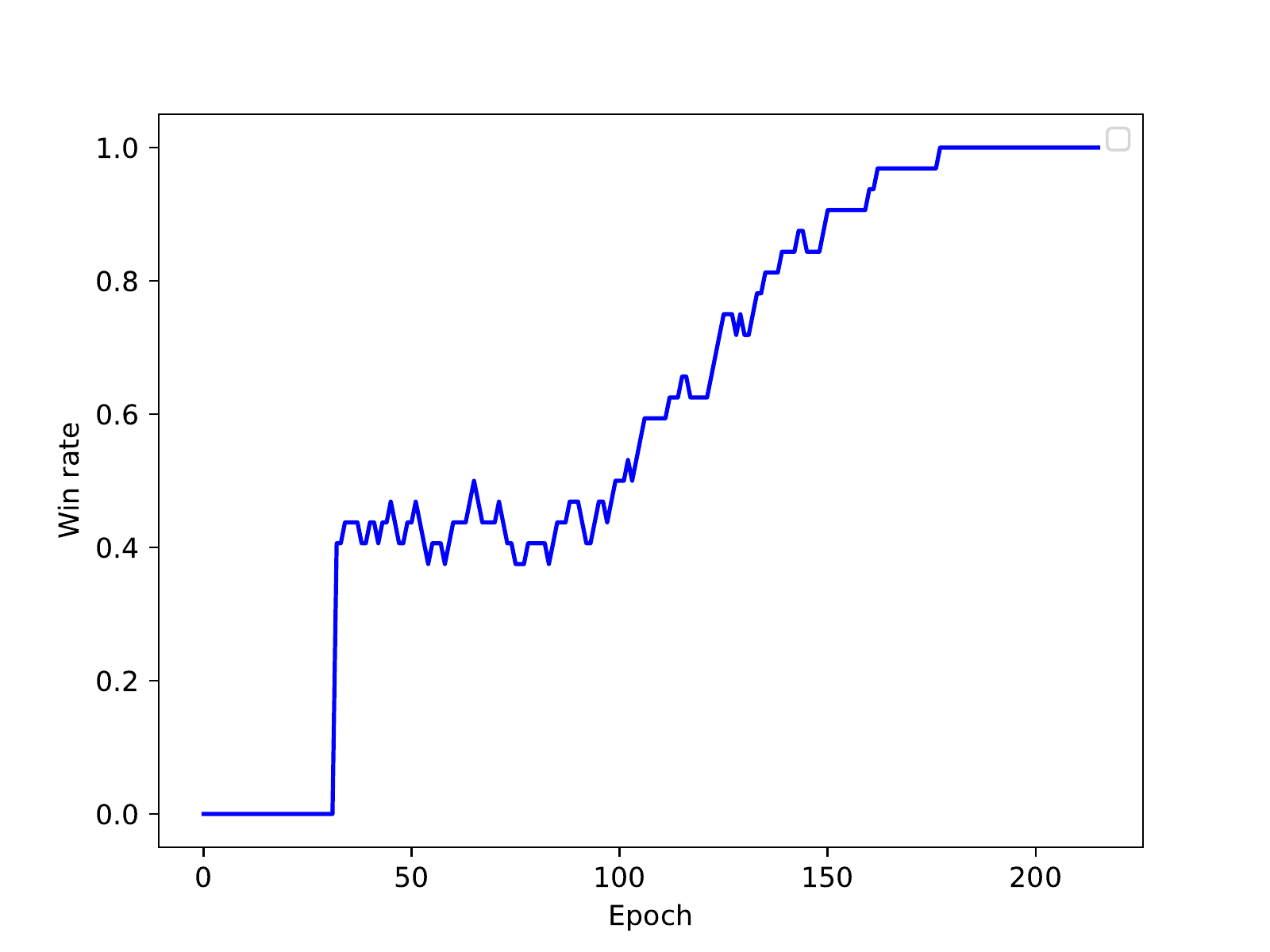}}
\caption{Navigation and obstacle avoidance results}
\label{fig-res-nav}
\end{figure}
\begin{figure}[!b]
\centering
\subfloat[Reward vs. episode graph\label{subfig-max-space-reward}]
{\includegraphics[width=0.47\linewidth]{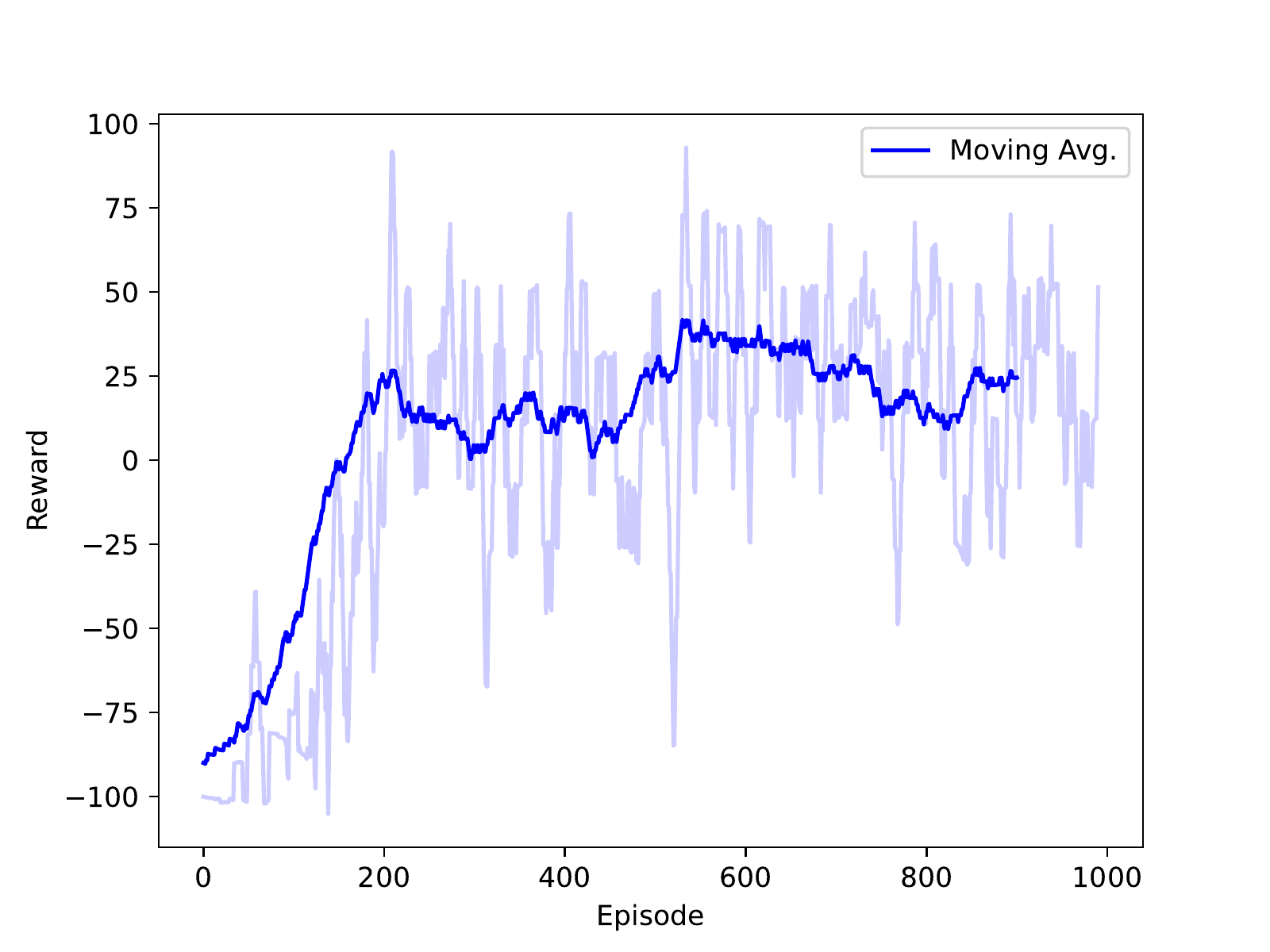}}
\hfil
\subfloat[Win rate vs. episode graph\label{subfig-max-space-rate}]
{\includegraphics[width=0.47\linewidth]{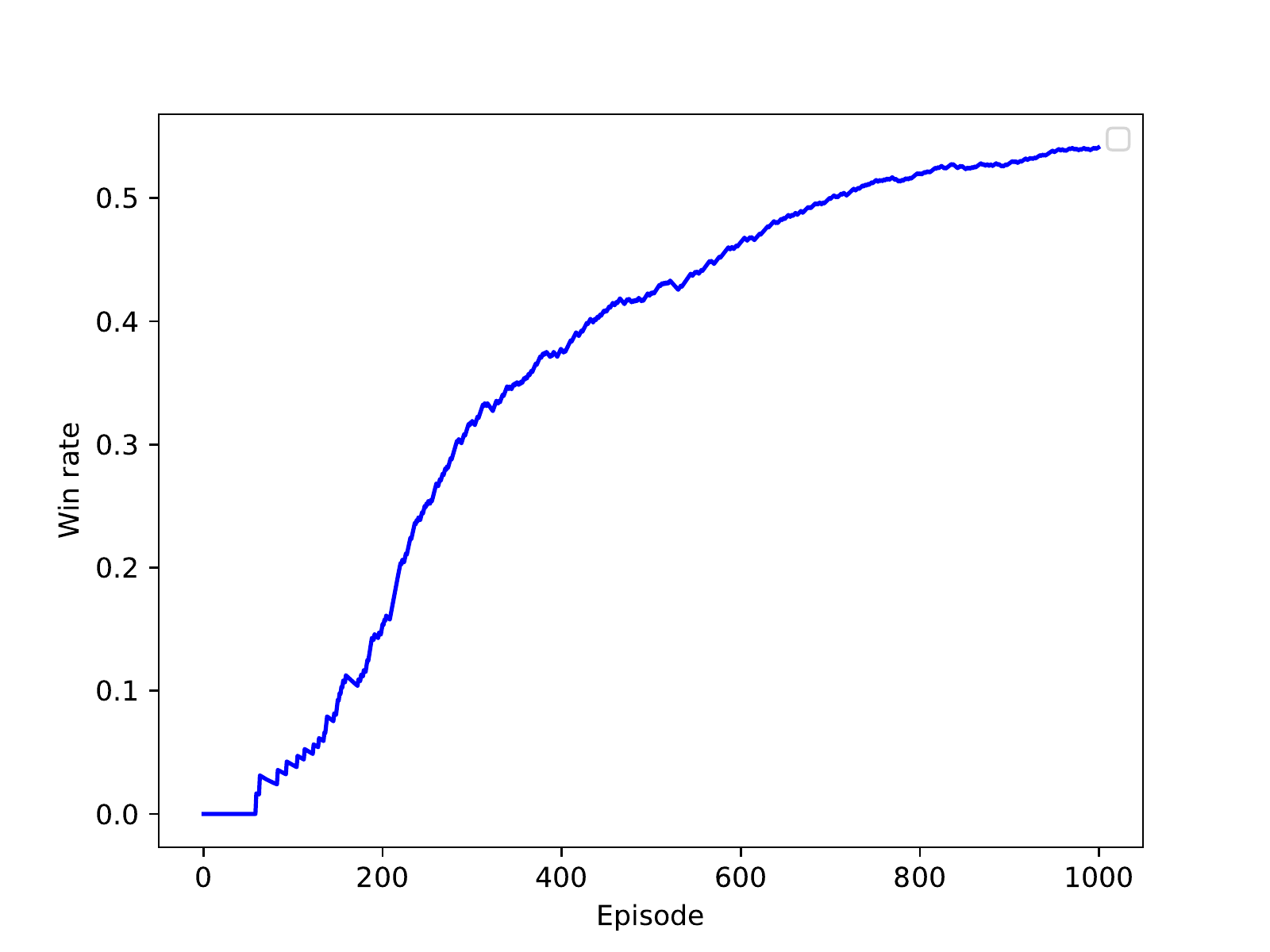}}
\caption{Maximum available space finding result}
\label{fig-res-max-space-both}
\end{figure}

Fig.~\ref{fig-res-nav} shows the training results of the navigation and obstacle avoidance model described in Section~\ref{sec-nav-obs}. During training, we determine that the training will occur up to 500 epochs. But if the agent can learn better policy before that, we stop the training process. We can call that situation an optimal policy when the agent gets a consistent win rate close to 1. In Fig.~\ref{fig-res-nav}, our agent learned a decent score between epoch numbers 200 to 220. Fig.~\ref{fig-res-nav}\subref{res-nav-1} represents the loss vs. epoch graph, and Fig.~\ref{fig-res-nav}\subref{res-nav-2} represents the win rate vs. epoch graph of our navigation and obstacle avoidance system during training. Fig.~\ref{fig-res-nav}\subref{res-nav-1} shows the line plot for the loss of the neural network during training. The light-blue line is the actual loss value, and the dark-blue line is the moving average of the loss value in this graph. The moving average is calculated according to Equation~(\ref{eq_matrix_moving_avg}). The line plot graph, especially the moving average plot, shows that the model is able to train the neural network so that the loss reduces gradually. Fig.~\ref{fig-res-nav}\subref{res-nav-2} is a line plot graph that shows that our model is becoming progressively better at reaching the destination without hitting anything. This graph shows that our agent is gradually increasing its winning rate to the point where the win rate becomes close to 1. 

\begin{equation}
    \label{eq_matrix_moving_avg}
    \text{Moving Average} (k, n) = \frac{\text{$\displaystyle\sum_{i=k}^{k+n} v_i $ }}{n}
\end{equation}


The training result in the maximum space finding model described in Section~\ref{sec-max-space} is shown in Fig.~\ref{fig-res-max-space-both}. We trained our agent for 1000 episodes. Fig.~\ref{fig-res-max-space-both} indicates that the agent successfully learned a better policy by gradually increasing the scores during training. Fig.~\ref{fig-res-max-space-both}\subref{subfig-max-space-reward} represents the reward vs. episode graph and Fig.~\ref{fig-res-max-space-both}\subref{subfig-max-space-rate} represents the win rate vs. episode graph of our maximum available space model. Fig.~\ref{fig-res-max-space-both}\subref{subfig-max-space-reward} shows the line plot graph of  gained rewards by the agent while taking action during training. The light-blue line is the actual reward value, and the dark-blue line is the moving average (according to Equation (\ref{eq_matrix_moving_avg})) of the reward value in this graph. In Fig.~\ref{fig-res-max-space-both}\subref{subfig-max-space-reward}, it appears that the agent gradually increases the rewards (according to Table~\ref{env-cell-value-b}). Fig.~\ref{fig-res-max-space-both}\subref{subfig-max-space-rate} represents the line plot graph of win rate while training. Win rate is calculated according to Equation~\ref{eq-matrix-rate}. This win rate is continuously increasing in this line graph. This graph is evidence that the agent is gaining optimal policy. 

\begin{equation}
    \label{eq-matrix-rate}
    \text{Rate} (k) = \frac{\text{$\displaystyle\sum_{i=0}^{k} v_i $ }}{k}
\end{equation}


\begin{figure}[!t]
\centering
\subfloat[Win rate vs. episode graph\label{subfig-multi-win-rate}]
{\includegraphics[width=0.47\linewidth]{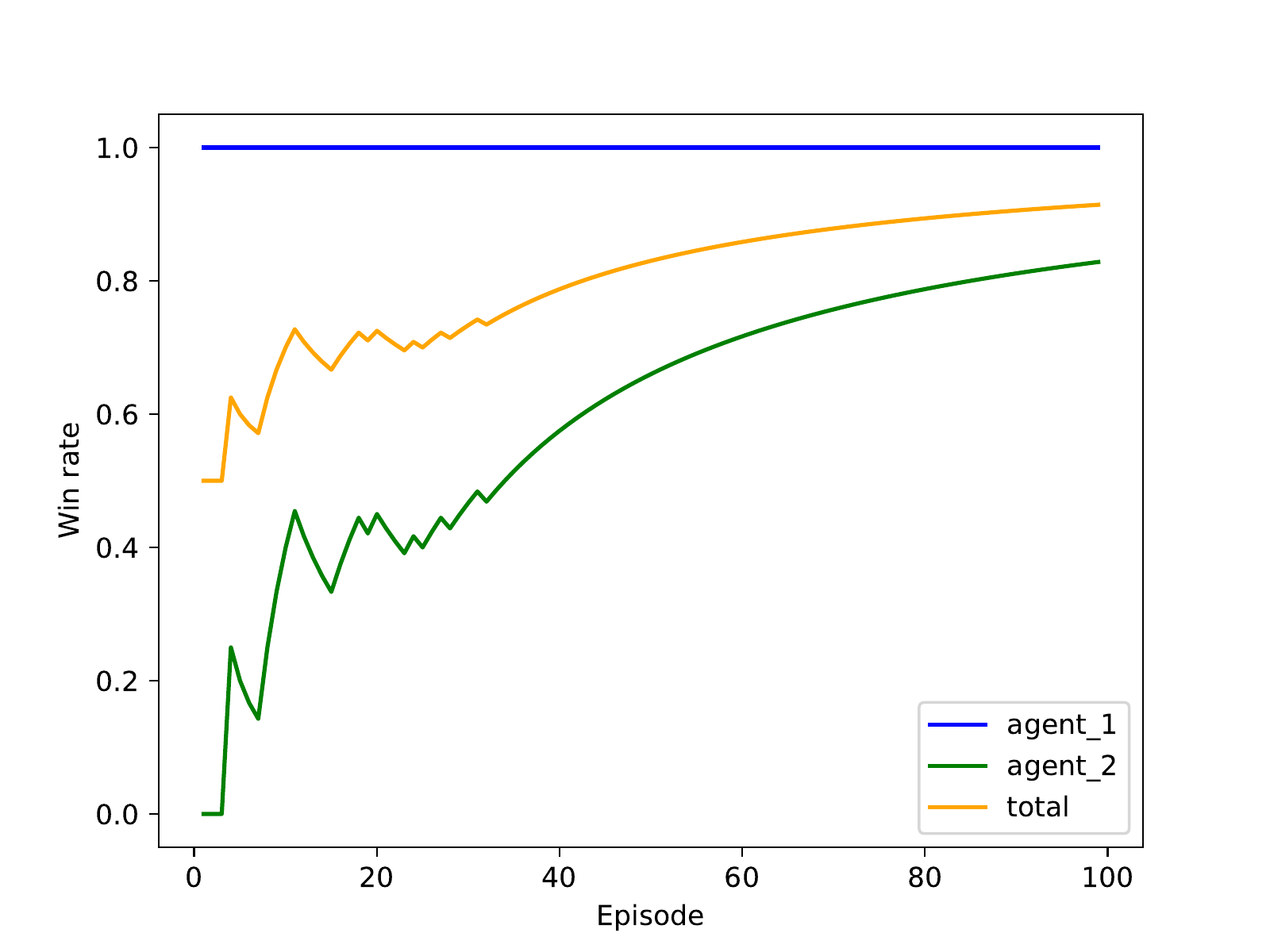}}
\hfil
\subfloat[Steps vs. episode graph\label{subfig-multi-steps}]
{\includegraphics[width=0.47\linewidth]{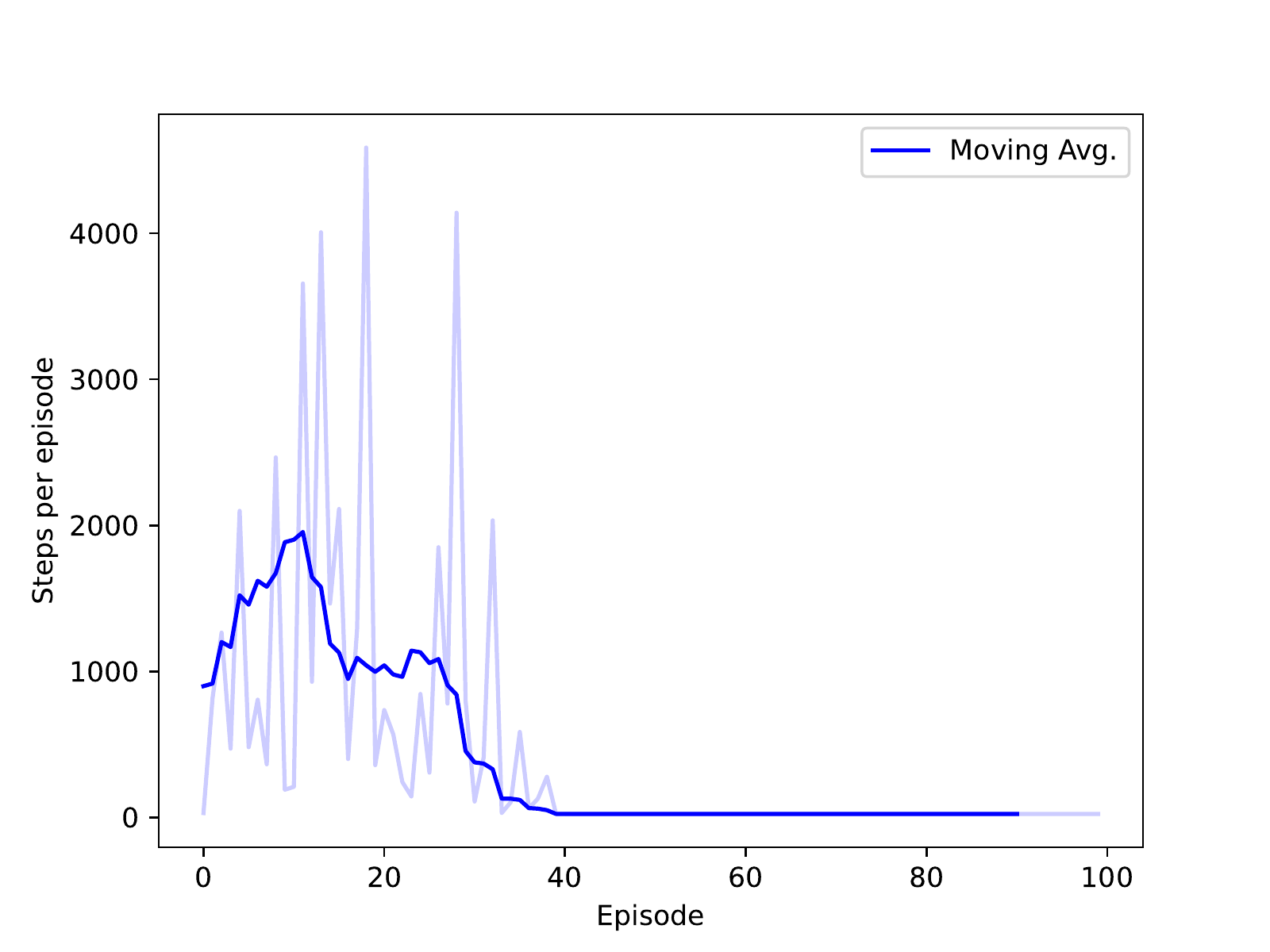}}
\caption{Multi-agent RL results}
\label{fig-res-multi-both}
\end{figure}

\begin{figure*}[!t]
\centering
\subfloat[Navigation and obstacle avoidance system\label{subfig-env-reach-nav-obs}]
{\includegraphics[width=0.16\linewidth, height=2.8cm]{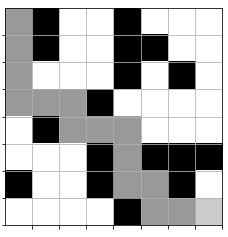}}
\hfil
\subfloat[Maximum available space finding\label{subfig-env-reach-max-space}]
{\includegraphics[width=0.16\linewidth]{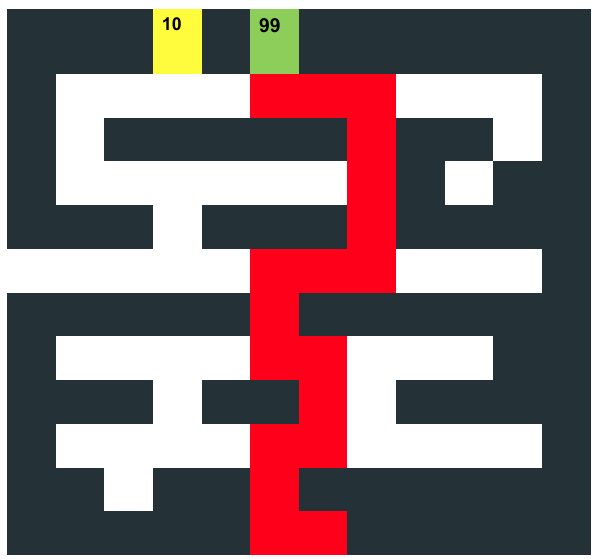}}
\hfil
\subfloat[Multi-agent environment\label{subfig-env-reach-multi}]
{\includegraphics[width=0.16\linewidth, height=2.5cm]{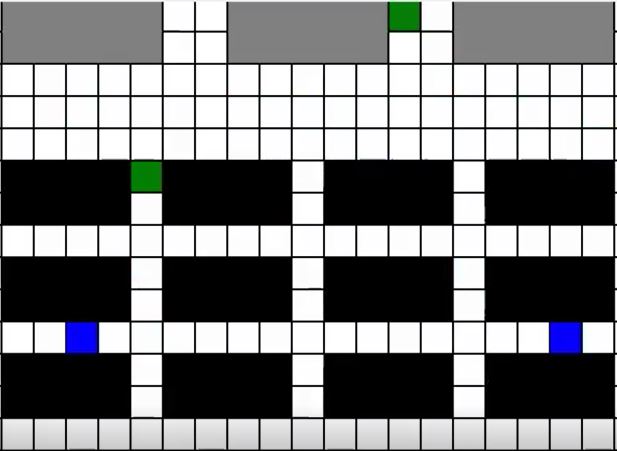}}
\caption{Visual demonstration of autonomous agents' navigation during the testing phase}
\label{env_reach_all}
\end{figure*}

The results of multi-agent RL expressed in Section~\ref{sec-multi-agent} are shown in Fig.~\ref{fig-res-multi-both}. We trained this multi-agent system with two agents for 100 episodes and recorded the results. Fig.~\ref{fig-res-multi-both}\subref{subfig-multi-win-rate} displays the line plot graph of win rate vs. episode for every acting agent, and Fig.~\ref{fig-res-multi-both}\subref{subfig-multi-steps} displays the graph for steps per episode vs. episode graph.  Fig.~\ref{fig-res-multi-both}\subref{subfig-multi-win-rate} is a multi-line plot graph that represents the win rate for two agents and the total rate. The win rate matrix is calculated by Equation~(\ref{eq-matrix-rate}). The three-line plots: blue, green, and orange constitute the win rate graph for the first agent, second agent, and total for both agents. Fig.~\ref{fig-res-multi-both}\subref{subfig-multi-steps} unveils the line plot for the steps needed for the agents to reach the destination points from starting points. The optimal model will take less time to reach the destination. The light-blue line draws the actual value, and the dark-blue line is the moving average value, which is calculated by Equation (\ref{eq_matrix_moving_avg}). This graph reveals that at first, the agents took many steps to reach the destination points, which is not convenient in these warehouse storing scenarios. But the agents gradually achieved a better policy to the point where they took the least number of steps to reach the destination because the line plot decreased afterward. 

We evaluate each of our models in their respective developed environments to observe the performance. Fig.~\ref{env_reach_all} unveils the visual representation of our agent navigating in the respective environments during the testing phase. Fig.~\ref{env_reach_all}\subref{subfig-env-reach-nav-obs} shows the path taken by the agent from starting point (upper-left) to destination point (lower-right). The agent's traversing area is the bold gray colored line. Fig.~\ref{env_reach_all}\subref{subfig-env-reach-max-space} displays the path taken by the agent described in Section~\ref{sec-max-space} during the testing phase. The red line is the path taken by the agent from starting point (lower-middle) to the destination point (upper-middle) containing maximum available space (100), which becomes 99 upon the agent's arrival. Finally, Fig.~\ref{env_reach_all}\subref{subfig-env-reach-multi} displays that the both agents (blue box) are at the destination point which is described Section~\ref{sec-multi-agent}. By observing these graphs, we can safely say that our three designed models can navigate and reach the destination points by following the shortest possible path, enabling our models to become time-efficient and resource-efficient.

\section{Conclusion}
\enlargethispage{-1.28in} 
In this paper, we design three approaches to navigate the autonomous robots in warehouse systems by using reinforcement learning. The first approach is designed with deep Q-learning, and the second one is developed with traditional Q-learning algorithms with slight variation. Both of these designs are for a single-agent environment. As we know that the practical usage of these autonomous systems will be in a multi-agent environment where optimal navigation and storage for the warehouse will take place, we design a multi-agent RL system for those scenarios. After that, we test and evaluate our designs' results and establish that all of our designs are suitable for use in practical fields as they unveil an excellent performance score for each type of warehouse environment. The results also establish that the autonomous agents reach the destination points by taking the least actions needed so that the cost of navigation remains low. The use of RL in a warehouse environment is ideal because the environment of these systems is dynamic, and RL is suitable to perform well in those partially observable, dynamic states. Although the use of RL algorithms in warehouse navigation is still moderate because of the lack of satisfactory design, we believe the use of RL algorithms in the design process will increase the possibility of deploying an autonomous system in real-world scenarios. In future work, we intend to design a multi-agent system that takes complex and higher dimensional inputs to classify and train the autonomous agents to deal with more practical scenarios ensuring that little or no intervention is needed once deployed.


\bibliographystyle{IEEEtran}

\bibliography{bibliography}

\end{document}